\documentclass{article}
\usepackage{natbib}
\usepackage[table]{xcolor}
\usepackage[utf8]{inputenc}
\usepackage[T1]{fontenc}
\usepackage{url}
\usepackage{booktabs}
\usepackage{amsfonts}
\usepackage{textalpha}
\usepackage{nicefrac}
\usepackage{microtype}
\usepackage{xcolor}
\usepackage{graphicx}
\usepackage{caption}
\usepackage{subcaption}
\usepackage{multirow}
\usepackage{makecell}
\usepackage{pifont}
\usepackage{enumitem}
\usepackage{amsmath}
\usepackage{amssymb}
\usepackage{wrapfig}
\usepackage{arydshln}
\usepackage{todonotes}
\usepackage{soul}
\setlist[enumerate]{leftmargin=*, label=(\arabic*), topsep=0pt}
\setlist[itemize]{leftmargin=*, topsep=0pt}
\usepackage{tikz}
\definecolor{yellowtext}{RGB}{68,132,243}
\definecolor{yellowred}{RGB}{50,167,82}
\definecolor{yellowblue}{RGB}{251,191,5}

\definecolor{darkblue}{rgb}{0, 0, 0.5}
\definecolor{chocolate}{HTML}{D2691E}
\definecolor{maroon}{HTML}{A00000}
\definecolor{indigo}{HTML}{4B0082}
\definecolor{violet}{HTML}{4B2E83}
\definecolor{lightblue}{rgb}{0.0, 0.0, 0.5}
\definecolor{cadmiumgreen}{rgb}{0.0, 0.42, 0.24}
\definecolor{forestgreen}{rgb}{0.13, 0.55, 0.13}
\definecolor{rowgray}{gray}{0.88}
\definecolor{sepgray}{gray}{0.55}
\definecolor{lightgrey}{gray}{0.92}
\definecolor{midgrey}{gray}{0.65}
\definecolor{headergrey}{gray}{0.30}
\usepackage[most]{tcolorbox}
\usepackage{fvextra}
\DefineVerbatimEnvironment{VerbatimWrap}{Verbatim}{
breaklines=true, 
breakindent=0pt, 
breaksymbol={},
fontsize=\scriptsize,
}

\usepackage[framemethod=tikz]{mdframed}
\mdfdefinestyle{customstyle}{
linecolor=cyan!70,
linewidth=3pt,
innerleftmargin=5pt,
topline=false,
rightline=false,
bottomline=false,
leftline=true,
innerrightmargin=5pt,
innertopmargin=5pt,
innerbottommargin=5pt,
backgroundcolor=cyan!8,
}

\usepackage{array}
\newcolumntype{P}[1]{>{\raggedright\arraybackslash}m{#1}}

\usepackage[colorlinks=true,linkcolor=blue,citecolor=green!50!black,urlcolor=magenta!70!black,filecolor=cyan!70!black]{hyperref}

\usepackage{pdflscape}
\usepackage{rotating}

\usepackage[table]{xcolor}
\definecolor{lightgray}{rgb}{0.9, 0.9, 0.9}

\usepackage{dcolumn}
\newcolumntype{C}{D{,}{,}{-1}} 
\newcolumntype{d}[1]{D{,}{,}{#1}}
\newcolumntype{.}{D{.}{.}{-1}}
\newcolumntype{,}{D{|}{|}{-1}}

\usepackage{lineno}

\definecolor{layer1}{HTML}{2B5797}    
\definecolor{layer2}{HTML}{2E7D32}    
\definecolor{layer3}{HTML}{E65100}    
\definecolor{auxprompt}{HTML}{616161} 

\newtcolorbox{promptbox}[2][]{%
  enhanced,
  breakable,
  colback  = white,
  colframe = #2!25,
  borderline west = {2.5pt}{0pt}{#2},
  title    = {#1},
  fonttitle = \bfseries\small,
  fontupper = \footnotesize,
  boxrule  = 0.3pt,
  left = 5pt, right = 5pt, top = 3pt, bottom = 3pt,
  before upper = {\setlength{\parskip}{4pt}\setlength{\parindent}{0pt}},
}

\usepackage{xspace}
\newcommand{\tool}{\textsc{TSAssistant}\xspace}
\newcommand{\tsa}{TSA\xspace}
\newcommand{\mcp}{MCP\xspace}
\newcommand{\hitl}{HITL\xspace}
\newcommand{\eg}{\textit{e.g.},\xspace}

\newcommand{\vs}{\textit{vs.}\xspace}

\usepackage{hyperref}

\usepackage[preprint]{icml2026}

\usepackage[capitalize,noabbrev]{cleveref}

\icmltitlerunning{\tool: Human-in-the-Loop Agentic Framework for Target Safety Assessment}
                  
\begin{document}

\twocolumn[
  \icmltitle{
    \tool: A Human-in-the-Loop Agentic Framework for Automated Target Safety Assessment
  }

  \icmlsetsymbol{equal}{*}
  \icmlsetsymbol{2nd}{$\dagger$}

  \begin{icmlauthorlist}
    \icmlauthor{Xiaochen Zheng}{equal,pm}
    \icmlauthor{Zhiwen Jiang}{equal,cscoe}
    \icmlauthor{David Tokar}{2nd,pm}
    \icmlauthor{Yexiang Cheng}{2nd,pm}
    \icmlauthor{Alvaro Serra}{pm}
    \icmlauthor{Melanie Guerard}{ts}
    \icmlauthor{Klas Hatje}{cscoe}
    \icmlauthor{Tatyana Doktorova}{pm}
  \end{icmlauthorlist}

  \icmlaffiliation{pm}{Predictive Modelling, Roche, Basel, Switzerland}
  \icmlaffiliation{cscoe}{Computational Sciences Center of Excellence, Roche, Basel, Switzerland}
  \icmlaffiliation{ts}{Translational Safety, Roche, Basel, Switzerland}

  \icmlcorrespondingauthor{Xiaochen Zheng}{xiaochen.zheng@roche.com}
  \icmlcorrespondingauthor{Zhiwen Jiang}{zhiwen.jiang@roche.com}

  \icmlkeywords{Target Safety Assessment, Multi-Agent Systems,
    Human-in-the-Loop, Drug Discovery,
    Large Language Models, Model Context Protocol,
    Retrieval-Augmented Generation}

  \vskip 0.3in
]

\printAffiliationsAndNotice{\icmlEqualContribution\icmlSecondEqualContribution}




\begin{abstract}

Target Safety Assessment (\tsa) requires systematic integration of genetic, transcriptomic, target homology, pharmacological, and clinical data to evaluate potential safety liabilities of therapeutic targets. This process is labor-intensive and expert-dependent, posing challenges in scalability and reproducibility. We present \tool, a human-in-the-loop multi-agent framework that decomposes \tsa report generation into a workflow of specialized subagents: \emph{Research Subagents} that each ground and cite a single \tsa domain, and \emph{Synthesis Subagents} that integrate findings across domains. Subagents retrieve and synthesize evidence from curated biomedical sources through standardized tool interfaces and produce individually citable, evidence-grounded sections, with behavior shaped by a hierarchical instruction architecture that separates coordination logic from domain expertise and user intent. To complement these soft constraints, programmatic execution hooks and persistent memory stores enforce hard constraints across the workflow, while an interactive refinement loop allows experts to review and revise individual sections with full conversational context preserved across iterations. Rather than a single holistic comparison, we decompose report quality into reproducibility, evidential grounding, task-level accuracy, and controllability under expert oversight, finding high reproducibility and grounding, substantial agreement with the human reference, and net-positive expert-driven refinement.

\end{abstract}

\section{Introduction}

\label{sec:intro}

Bringing a new medicine from target identification to regulatory approval typically spans more than a decade, with only a small fraction of candidates that enter clinical trials ultimately reaching patients~\citep{olson2000concordance, hay2014clinical, sun202290}. 
Among the root causes of late-stage failure, inadequate early characterization of target-related safety liabilities is consistently identified as a primary contributor~\citep{cook2014lessons, harrison2016phase}. 
Target Safety Assessment (\tsa) is the systematic process of evaluating whether pharmacological modulation of a biological target is likely to produce adverse on-target or off-target effects in humans. It also supports species selection for subsequent animal testing by assessing the similarity between human gene sequences and those of relevant preclinical species, and integrates insights from the competitive landscape for the same target, including known severe adverse events.
Performed at the target nomination stage and conducted rigorously, \tsa can de-risk programs before resources are committed to preclinical and clinical studies~\citep{plenge2013validating}.

Despite its importance, the generation of \tsa reports remains largely manual, with content composition not standardized and the level of detail and depth of analysis often varying between individuals. Additionally, scientists must synthesize heterogeneous evidence spanning human genetics~\citep{nelson2015support,
buniello2019nhgri}, tissue-expression profiles~\citep{uhlen2015tissue, gtex2020gtex}, pharmacological interaction data~\citep{wishart2018drugbank, mendez2019chembl}, and adverse-event repositories~\citep{pinero2020disgenet, banda2016curated}.
This process is time-consuming, difficult to harmonize and communicate across teams, and prone to knowledge gaps when evidence bases grow and portfolios scale~\citep{morgan2018impact}.

Recent advances in large language model (LLM) agents offer a promising route to augmenting this workflow. While general-purpose LLMs (\eg GPT, Gemini) can be applied through prompt-based approaches, such usage alone often lacks consistency and reproducibility. Tool-augmented LLMs~\citep{bran2024chemcrow, boiko2023autonomous}, multi-agent coordination frameworks~\citep{wu2023autogen, hong2024metagpt}, and retrieval-augmented generation (RAG)~\citep{lewis2020retrieval} address these limitations by introducing structured, agentic coordination, and have demonstrated that complex, multi-step scientific reasoning is achievable at scale. Autonomous scientific systems such as the AI Co-Scientist~\citep{gottweis2025aicoscientist} and the AI Scientist~\citep{lu2024aiscientist} further demonstrate the potential of agent-based approaches for knowledge-intensive research tasks. 
However, pharmaceutical workflows with safety endpoints require more than automation: key requirements include human oversight, transparent evidence traceability, and fine-grained expert control~\citep{amershi2014power, mosqueira2023human}.

We introduce \tool, a human-in-the-loop multi-agent framework for \tsa. Our principal contributions are: \textbf{(i)} a \textbf{section-based multi-agent workflow} in which specialized subagents independently ground and cite each \tsa domain (\cref{fig:hierarchy}); \textbf{(ii)} a \textbf{three-layer instruction architecture} separating coordination logic, domain skills, and runtime user intent; \textbf{(iii)} a \textbf{programmatic enforcement layer} comprising execution hooks and persistent memory stores, together with domain-engineered tool interfaces that encapsulate multi-step expert analytical workflows rather than serving as simple API wrappers; and \textbf{(iv)} an \textbf{interactive refinement loop} with tool and agent memory, allowing section-level targeted revision and expert-guided iteration (\cref{fig:loop}).

\section{Background}
\label{sec:background}

\subsection{Target Safety Assessment}
\label{sec:bg:tsa}

\tsa is conducted at the target nomination stage in drug discovery to evaluate potential adverse effects of modulating a biological target in humans, while informing species selection and integrating competitive landscape insights to de-risk programs before preclinical and clinical development~\citep{cook2014lessons}. A comprehensive assessment typically integrates multiple lines of evidence, grounded in an understanding of target biology, including: 
\textbf{(i)}~\textit{Genetic evidence}, human genetic variants linking target perturbation to phenotypic consequences~\citep{plenge2013validating, nelson2015support, buniello2019nhgri}; 
\textbf{(ii)}~\textit{Transcriptomic evidence}, tissue- and cell-type-specific expression profiles to identify undesired expression in off-target tissues ~\citep{uhlen2015tissue, gtex2020gtex}; 
\textbf{(iii)}~\textit{Target homology}, sequence and structural similarity to proteins that could produce off-target pharmacology~\citep{altschul1997gapped, jumper2021alphafold}; \textbf{(iv)}~\textit{Pharmacological evidence}, known effects of approved or investigational drugs modulating the same target~\citep{wishart2018drugbank, mendez2019chembl, santos2017comprehensive}; and 
\textbf{(v)}~\textit{Clinical evidence}, signals of adverse events and disease associations from clinical registries and databases~\citep{pinero2020disgenet, ochoa2021open}.
Structured frameworks such as the AstraZeneca 5R guidelines~\citep{cook2014lessons} have codified these requirements, yet execution remains inconsistently documented, expert-dependent, and difficult to scale across large target portfolios.

\subsection{LLM-Based Scientific Multi-Agent Systems}
\label{sec:bg:agents}

LLMs have evolved from single-pass text generators into systems capable of multi-step reasoning, tool use, and interaction with external knowledge sources~\citep{wei2022chain, yao2022react, lewis2020retrieval}, enabling agentic frameworks in which specialized agents collaborate on multi-step tasks~\citep{chen2024reconcile, kim2024mdagents, zhou2025mam, wang2025colacare, zhang2026virtual, zhu2026medagentboard}.

\textbf{Multi-Agent Decomposition and Memory.} A single agent's context window must simultaneously manage task planning, domain knowledge, tool interactions, and output formatting, which creates capacity bottlenecks and makes individual concerns difficult to maintain independently~\citep{yao2022react, shinn2023reflexion, yao2023tree}. Multi-agent systems mitigate this by decomposing complex tasks across subagents, enabling modular distribution and specialization~\citep{wu2023autogen, hong2024metagpt}. In more structured systems, the concerns governing each agent (its role, coordination protocol, and domain knowledge) are factored into separate components rather than fused into a single specification. \tool inherits this separation of concerns and extends it to the instruction level through its three-layer architecture (\cref{sec:method:instructions}).

A complementary challenge is maintaining coherence across agents without exceeding context limits. Generative Agents~\citep{park2023generative} address this through a persistent external memory stream from which relevant observations are retrieved for injection into the current agent context, decoupling memory persistence from the LLM's ephemeral context window. \tool instantiates this principle through its tool memory and agent memory adapted to a structured workflow (\cref{sec:method:enforcement}). 

\textbf{Autonomous Scientific Agents.} Recent work extends multi-agent architectures to full scientific workflows. The AI Scientist~\citep{lu2024aiscientist} automates the research cycle from idea generation to manuscript preparation, and the AI Co-Scientist~\citep{gottweis2025aicoscientist} uses tournament-based generation, reflection, and ranking to refine biomedical hypotheses. The Virtual Lab~\citep{swanson2025virtual} couples computational design with experimental validation, while ToolUniverse~\citep{gao2025democratizing} standardizes tool interfaces for non-expert users.

\tool differs in several key respects. It targets structured evidence synthesis over a pre-defined report schema rather than open-ended exploration; it operates in pre-clinical safety assessment, where traceability, reproducibility, and human accountability are primary requirements; and it embeds expert-defined constraints and decision frameworks to replicate the workflow of target safety experts. These requirements motivate the programmatic enforcement layer and section-level human-in-the-loop (\hitl) design, as described in \cref{sec:method}.

\subsection{Human-in-the-Loop Systems}
\label{sec:bg:hitl}

\hitl systems integrate human judgment into automated workflows and decision processes to improve accuracy, safety, and user trust, especially in high-risk and regulated domains~\citep{amershi2014power, mosqueira2023human}. Rather than treating models as fully autonomous, \hitl systems enable iterative interaction in which users can guide, correct, and validate intermediate outputs. 

In pharmaceutical research and development, regulatory guidance explicitly requires expert accountability for safety decisions, making full automation structurally inappropriate regardless of performance~\citep{cook2014lessons}. Existing guidance emphasizes that automated systems must support, rather than replace, human judgment. We apply this constraint to \tsa, where erroneous assessments carry downstream consequences for patient safety and research resource allocation. \tool instantiates these principles through a section-level \hitl design described in \cref{sec:method:loop}.


\section{\tool: System Design}
\label{sec:method}

\begin{figure*}[ht]
  \centering
  \includegraphics[width=0.95\textwidth]{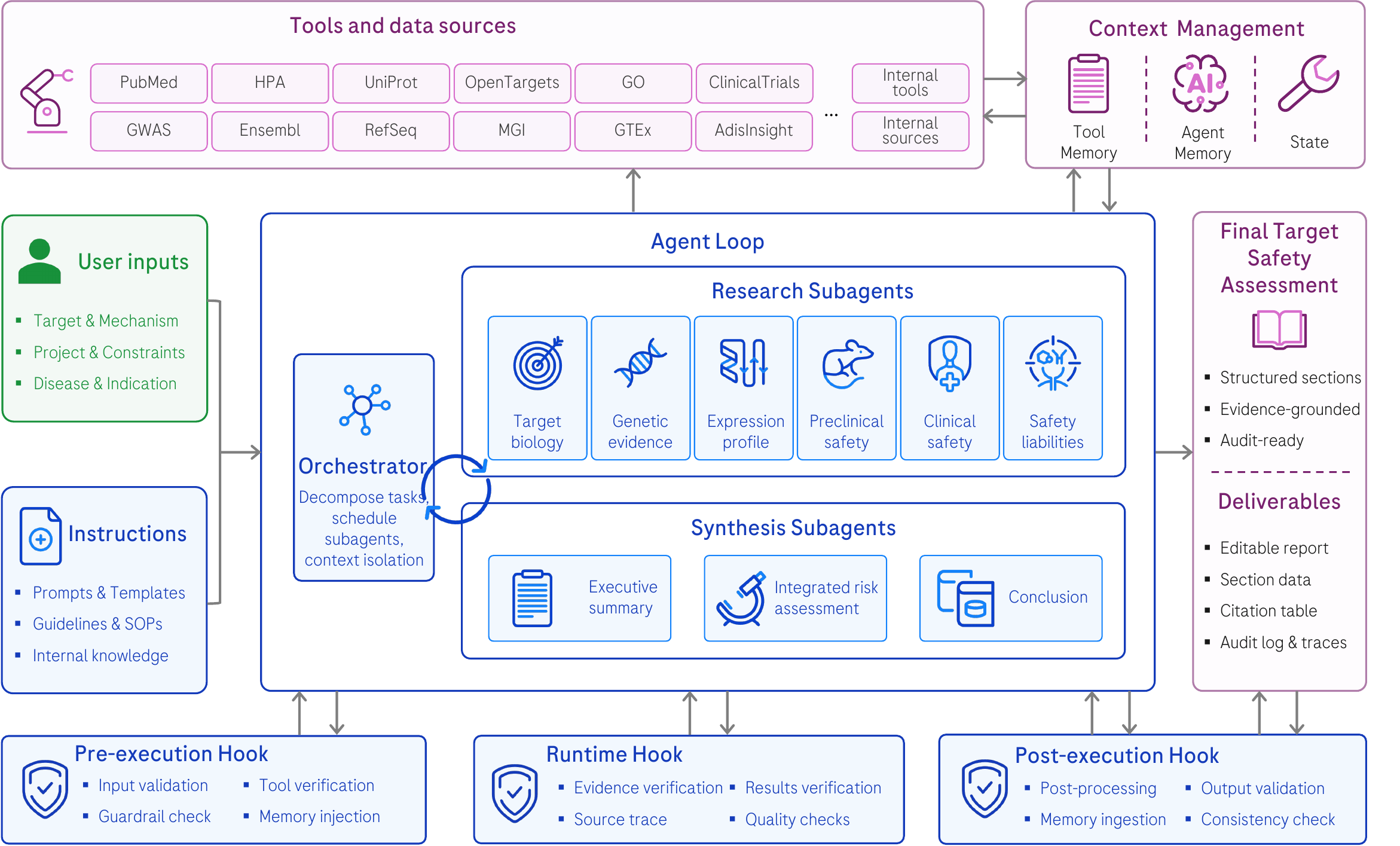}
  \caption{
    \textbf{Hierarchical agent architecture of \tool.} An \emph{Orchestrator} decomposes the assessment into \emph{Research Subagents}, each covering one \tsa domain, and \emph{Synthesis Subagents} that integrate findings across domains. Execution hooks and three persistent context stores (tool memory, agent memory, execution state) provide programmatic enforcement; all subagents access curated biomedical data sources through standardized \mcp tool interfaces. Mechanism detail in \cref{sec:method:enforcement}.
  }
  \label{fig:hierarchy}
\end{figure*}
Given a biological target identifier (\eg gene ID or symbol, UniProt accession) and optional project context including disease therapeutic area and compound modality, \tool produces a structured \tsa report comprising individually citable, evidence-grounded sections. The framework decomposes report generation into a workflow of specialized subagents: \emph{Research Subagents}, each responsible for one \tsa domain, followed by \emph{Synthesis Subagents} that integrate findings across domains (\cref{fig:hierarchy}). Each domain draws on one or more of the five lines of evidence (\cref{sec:bg:tsa}) rather than mapping to them one-to-one, and the Synthesis Subagents integrate across completed domains without issuing new retrievals.

\begin{figure*}[t]
  \centering
  \includegraphics[width=0.95\textwidth]{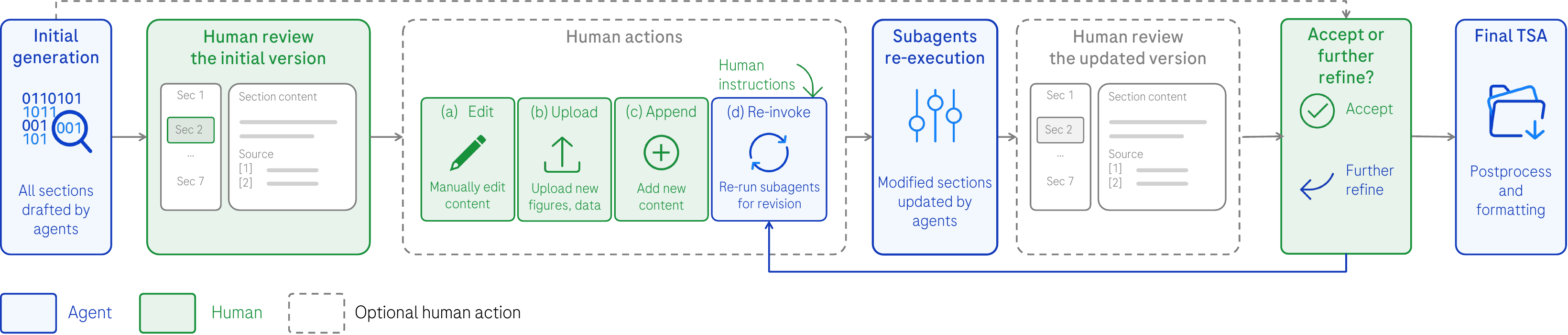}
  \caption{
    \textbf{Interactive refinement loop in \tool.}
    After initial section generation, the user reviews each section and may: \textbf{(a)} manually edit content, \textbf{(b)} append new information, \textbf{(c)} upload additional sources or graphics, or \textbf{(d)} re-invoke the subagent for targeted revision. Conversational memory preserves context across iterations; user feedback progressively adapts retrieval strategies and prompt templates.
  }
  \label{fig:loop}
\end{figure*}

A central design challenge is that LLM agent behavior is governed by prompts, which function as \emph{soft constraints}: they bias model outputs statistically but do not reliably guarantee correctness, reproducibility, or evidence fidelity. In monolithic single-agent approaches, errors that enter the shared context risk compounding across the report, and multi-agent decomposition alone does not resolve this without explicit verification mechanisms~\citep{kim2025towards}. These observations motivate a complementary architecture in which a \emph{hierarchical instruction architecture} (\cref{sec:method:instructions}) organizes agent behavior through layered prompts, a \emph{programmatic enforcement layer} (\cref{sec:method:enforcement}) of tool interfaces, execution hooks, and context management provides hard constraints, and an \emph{interactive refinement loop} (\cref{sec:method:loop}, \cref{fig:loop}) embeds expert oversight at the section level.

\subsection{Hierarchical Instruction Architecture}
\label{sec:method:instructions}

Agent behavior in \tool is governed by a three-layer instruction hierarchy that applies \emph{separation of concerns} to prompt engineering~\citep{hong2024metagpt, wu2023autogen}, factoring instructions by change rate and stakeholder.

\textbf{Layer~1: System Prompts.} Define agent roles, coordination logic, and output structure, shared immutably across all targets.

\textbf{Layer~2: Domain Skill Modules.} Encode section-specific research and tool-use strategies, evidence weighting, and writing guidelines as independently versioned, composable modules.

\textbf{Layer~3: Runtime User Instructions.} Inject project-specific constraints at highest priority, overriding lower-layer defaults.

\noindent At runtime the layers compose with higher layers taking precedence, allowing domain experts to evolve research guidance independently of coordination logic, and users to customize individual assessments without modifying either reusable layer.

\subsection{Programmatic Enforcement Layer}
\label{sec:method:enforcement}

\textbf{Tool Interfaces.} Each subagent operates through \mcp-standardized tool interfaces~\citep{anthropic2024mcp} that implement complete analytical subroutines rather than serving as data-retrieval endpoints alone. The guiding principle is to decompose expert research workflows into atomic, deterministic operations implemented in auditable code: the tool layer fixes each analytical procedure (data acquisition, computational analysis, rule-based interpretation) in code rather than delegating it to the LLM's stochastic interpretation of raw data. Tool pipelines span multiple levels of abstraction, from low-level operations such as ontology traversal, sequence-level computation, and statistical aggregation, to higher-level domain logic encoding acceptance criteria and multi-source cross-referencing strategies that would otherwise demand expert judgment at each invocation. Every returned record is automatically indexed in the tool's paired memory store with full provenance metadata, enabling systematic comparison against expert reference standards (\cref{sec:experiments}). The framework additionally integrates internal proprietary tools, withheld for confidentiality. A complete catalog of the public data sources backing these tools is provided in \cref{app:datasources}.

\textbf{Execution Hooks.} Execution hooks wrap each subagent's lifecycle (\cref{fig:hierarchy}): before execution, they inject compressed upstream findings and serialize subagent launches; after execution, they validate inline citations against retrieved evidence via natural-language inference and compress the output for downstream consumption; during execution, they cross-reference each newly retrieved record against accumulated evidence to detect contradictions.

\textbf{Context Management.} Each subagent executes in an isolated context to prevent cross-contamination and context window overflow. Three persistent stores bridge the isolated agents.

\emph{Tool memory} records every tool-call output with provenance metadata (invoking subagent, tool, query, pipeline stage), each assigned a globally unique identifier with full CRUD support, enabling downstream lookup and cross-referencing without re-execution.

\emph{Agent memory} stores compressed factual summaries of completed sections and selectively injects them into downstream subagents via a predefined dependency graph.

\emph{Execution state} is persisted for resumability: an interrupted run resumes from the last completed section, preserving accumulated tool and agent memory.

\subsection{Human-in-the-Loop Refinement}
\label{sec:method:loop}

After initial report generation, \tool enters an interactive refinement loop (\cref{fig:loop}) in which experts and the system iterate on the draft. Users can edit section content directly, append new information, upload additional sources, or re-invoke the responsible subagent to revise a specific section. A conversational memory store carries context across iterations, so users do not need to restate background on each interaction, and captured corrections progressively adapt retrieval constraints and prompt templates to team conventions and project context.

A deliberate design choice is that human validation occurs at the \emph{section} level rather than only at report completion. This enables early error detection and reduces the cost of late corrections, since a flawed upstream section (\eg an incorrect knockout phenotype) would otherwise propagate into downstream synthesis sections before the expert has an opportunity to intervene.

More broadly, \tool adopts a hybrid mode in which the LLM handles research and drafting while domain experts retain final decision authority~\citep{amershi2014power, mosqueira2023human}. This division reflects both regulatory requirements for expert accountability in safety decisions and the practical limitations of current LLMs in resolving ambiguous evidence~\citep{gabriel2024ethics}.

\section{Evaluation Framework}
\label{sec:experiments}

Comprehensive evaluation of \tsa report quality can be approached from two directions: externally, by comparing system outputs against expert-authored reference reports across dimensions such as factual consistency, evidence completeness, and structural alignment; and intrinsically, by measuring properties of the generated report itself, including self-consistency across independent runs and traceable grounding of individual claims. External evaluation is, however, inherently difficult to scale: evidence selection and interpretation involve expert judgment, and two independently authored reports on the same target may legitimately differ in scope and emphasis while both being scientifically valid.

All system evaluations are conducted across 35 drug targets spanning diverse therapeutic areas and mechanism classes; the sequence homology benchmark uses a 22-target subset for which expert-authored reference analyses are available.
We evaluate \tool along two complementary axes. \emph{Automatic intrinsic} evaluations (\cref{sec:eval:auto}) probe report-level reliability without any human reference, through claim-level self-consistency across independent runs (\cref{sec:eval:concordance}) and sentence-level evidential grounding (\cref{sec:eval:esr}). \emph{Evaluations with human reference and oversight} (\cref{sec:eval:human}) capture the two ways human expertise enters the loop: a task-level benchmark (\cref{sec:eval:seqhom}) compares the agent's ortholog selection against analyses performed by human computational biologists as a frozen reference, and an expert-in-the-loop refinement test (\cref{sec:eval:hitl}) measures whether interactive expert oversight produces precise, beneficial revisions or introduces collateral damage. Together, our evaluation strategy decomposes report quality into independently measurable properties: reproducibility, evidential grounding, task-level accuracy, and controllability under expert oversight.

\subsection{Automatic Intrinsic Evaluation}
\label{sec:eval:auto}

\subsubsection{Claim-Level Self-Consistency}
\label{sec:eval:concordance}
The integrated risk assessment produced by \tool consolidates findings into a structured table of 15 standardized toxicity endpoints (\cref{tab:endpoints} in \cref{app:eval_defs}) aligned to MedDRA System Organ Classes~\citep{brown1999medical}.
We formalize self-consistency~\citep{manakul2023selfcheckgpt, kuhn2023semantic, lin2024generating} measurement as a \emph{structured concordance problem}: given two independent runs on the same target, how well do they agree on the specific evidentiary claims supporting each safety endpoint?
Holistic text similarity fails to capture this, as two reports may use different phrasing for the same observation or cite the same phenotype from different biological contexts.

Following FActScore~\citep{min2023factscore} and SAFE~\citep{wei2024longform}, we decompose each endpoint's findings into structured atomic claims\footnote{In the safety and toxicology context, each atomic claim corresponds to a single empirical observation: either a specific preclinical phenotype (\eg a knockout mouse phenotype or histopathological finding) or a clinical adverse event (\eg an SAE or TEAE from a clinical trial). We retain the term ``claim'' to align with the NLP evaluation literature.} $c = (\textit{observation}, \textit{species})$, where \textit{species} encodes the biological context in which the observation was made.
Unlike prior work where each atomic fact is a plain-text statement, toxicological evidence is inherently structured: the same phenotype observed in different species carries different translatability for human risk.
Pairs of claims from two independent runs are scored by an LLM-as-judge binary semantic similarity~\citep{zheng2023judging, liu2023geval, jiang2025clear}, scaled by a species-context weight $\phi_s$ reflecting biological translatability informed by regulatory toxicology practice~\citep{ich2011s6} (same species: 1.0, same species group: 0.75, cross-group: 0.5). Scores are aggregated via symmetric greedy matching~\citep{zhang2020bertscore} into Precision, Recall, and Claim-F1, where neither run is treated as ground truth (full metric definitions in \cref{app:concordance_method}).

\paragraph{Experimental setup.} 
We evaluate across 35 drug targets, each assessed with 10 independent \tool-generated reports.
For each target, the integrated risk assessment table is parsed to extract the structured endpoint findings, aligned to the 15 predefined endpoints.
Claim extraction uses Claude~Opus~4.6; similarity assessment uses Claude~Opus~4.6~\footnote{A single pairwise LLM-as-a-judge evaluation requires $35 \times 45 \times 15 \times \bar{n}$ API calls, where $\bar{n}$ is the average number of claim pairs per endpoint ($\bar{n} \approx 10$, range $[0, 20]$), i.e., roughly $2.3 \times 10^5$ calls. We therefore restrict our experiments to a single judge model and leave cross-LLM comparison to future work.}.
Claim pairs that fail the species context filter ($\phi_s = 0$) are excluded from the matching rather than scored as zero, as direct translatability between these evidence types cannot be assumed.
We score all $\binom{10}{2} = 45$ pairs of \tool runs per target using an all-pairs design rather than designating a single run as ground truth~\citep{manakul2023selfcheckgpt, kuhn2023semantic}, yielding a symmetric self-consistency estimate.

\begin{figure}[t]
  \centering
  \includegraphics[width=0.9\linewidth]{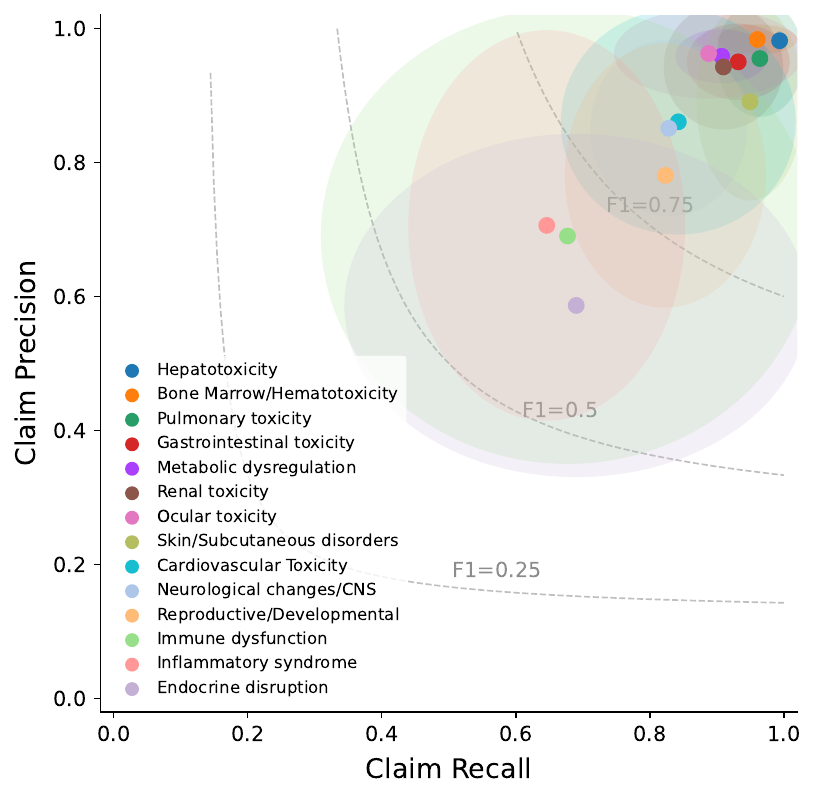}
  \caption{
    \textbf{Claim-Level Self-Consistency by Safety Endpoint.} Symmetric Precision and Recall (\cref{sec:eval:concordance}) for each of the 15 endpoints, aggregated over 45 run pairs per target across 35 targets. 
    The shaded ellipsoidal regions around each endpoint are axis-aligned and span $\pm 1$ sample standard deviation of Precision and Recall computed across all $35 \times 45$ (target, run-pair) observations per endpoint. Tighter ellipses indicate that an endpoint's self-consistency is stable across both runs and biological contexts; broader ellipses indicate substantial variation, typically correlating with evidence sparsity or domain-specific phenotype ambiguity. F1 contours are shown at $0.25$, $0.50$, and $0.75$.
  }
  \label{fig:claim_pr}
\end{figure}

\paragraph{Results.}
\Cref{fig:claim_pr} reports claim-level self-consistency across the 15 safety endpoints. Because neither run is treated as ground truth, Precision and Recall here quantify how many claims in one run have a species-compatible match in the other and vice versa. \Cref{tab:species_relation_rules} further breaks down all the atomic claim pairs by species relation and scoring rule.

Most of the endpoints achieve both Precision and Recall above $0.75$, with Claim-F1 scores above the $0.75$ contour, supporting the conclusion that \tool produces reproducible endpoint-level findings across independent runs. 
Several endpoints, including hepatotoxicity, hematotoxicity, pulmonary toxicity, and gastrointestinal toxicity, approach Claim-F1 values close to $1.0$, consistent with the structured and well-characterized nature of evidence sources for these endpoints, indicating that independent runs converge on the same atomic claims with matching species context across diverse targets.

A smaller number of endpoints, particularly endocrine disruption, inflammatory syndrome, immune dysfunction, exhibit lower Claim-F1 driven by reduced Precision and/or Recall. We attribute this to greater mechanistic heterogeneity in the underlying evidence base, sparser per-target data density, and increased ambiguity in phenotype terminology and adverse event classification in these domains, all of which broaden the space of plausible findings any single run may surface and therefore reduce claim-level overlap between independent runs. Additional retrieval guidance, refined skill modules, or more targeted human review in the refinement loop (\cref{sec:method:loop}) are most warranted in subsequent iterations.

\subsubsection{Sentence-Level Grounding}
\label{sec:eval:esr}

To evaluate the factual reliability of generated \tsa reports, we introduce the \textit{evidential support rate} (ESR), which measures the proportion of factual sentences supported by traceable evidence, serving as an inverse indicator of hallucination.

\paragraph{Metric.}
Each sentence in a report is classified by an LLM judge (Claude~Opus~4.6) into one of seven categories: \textit{cited} (has an inline reference), \textit{tool-grounded} (verified by MCP tool call results), \textit{inferable} (derivable from preceding grounded context), \textit{common knowledge}, \textit{procedural} (methodology or guidance), \textit{in-memory} (supported by subagent memory from prior sections), or \textit{unsupported} (an ungrounded factual claim and potential hallucination). The ESR is then computed as the fraction of factual sentences (excluding procedural classifications) that are not unsupported.

\paragraph{Experimental setup.}

\begin{figure}[t]
    \centering
    \includegraphics[width=\linewidth]{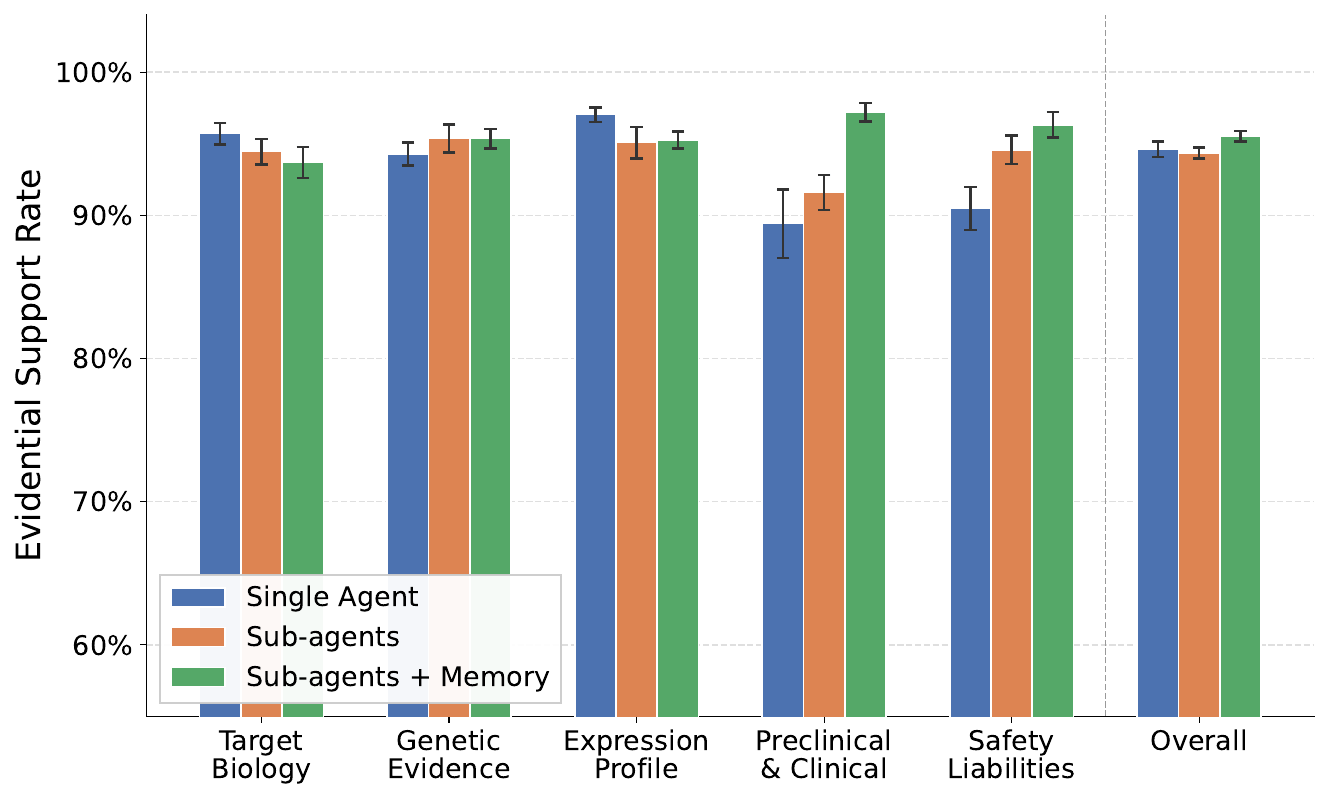}
    \caption{Evidential support rate by section. Error bars indicate standard error.}
    \label{fig:esr}
\end{figure}

\Cref{fig:esr} compares the ESR across three architectural conditions (single agent, subagents, and subagents with memory), averaged over 35 drug targets.

\paragraph{Results.}

The single agent scores highest on early sections (Target Biology, Expression Profile), where its context window is still short and manageable. As it generates more sections and the accumulated context grows, its ESR drops to $\sim$90\% on Preclinical \& Clinical and Safety Liabilities, suggesting that the model struggles to faithfully attend to relevant information buried in a long conversation history. Subagents sidestep this issue by starting each section with a fresh context. With memory enabled, downstream subagents receive prior findings directly, allowing them to cite or build on earlier evidence. This combination yields the highest overall ESR and reduces the late-section degradation observed in the single-agent condition, consistent with both task decomposition and cross-section evidence propagation contributing to grounding in \tsa report generation.

\subsection{Evaluation with Human Reference and Oversight}
\label{sec:eval:human}

\subsubsection{Task-Level Benchmark: Sequence Homology}
\label{sec:eval:seqhom}

Target homology is one of the five lines of evidence of \tsa (\cref{sec:bg:tsa}), and its computational core is ortholog selection: for each preclinical species, identify the protein sequence most likely to recapitulate the target's biology. We benchmark the agent's ortholog selection against expert-curated ground truth from 22 \tsa programs, asking whether it is reliable enough to serve as the homology subagent in deployment.

The task spans 14 ortholog databases with biological and database-driven edge cases (divergent isoforms, long lineage-specific extensions, partial annotations). The agent works within a dedicated memory environment that references sequences by protein ID rather than transcribing full sequences, and applies a deterministic pre-filter that discards candidates differing from the human canonical sequence by more than $20\%$ in length or showing more than a $60\%$ mismatch in pairwise alignment. When no suitable candidate is found, the agent excludes the species and reports every database it searched together with the rejection reason, so the user can override (full procedure in \cref{app:seqhom_procedure}).

For each species both human and agent included in their final results, we compare protein sequences of the orthologs by pairwise alignment rather than by identifier (the same sequence may be retrieved from different databases). Based on the percent sequence identity of this alignment, we classify each comparison into three tiers: an \emph{exact match} ($100\%$ identity), where the agent and human selected the same protein sequence; a \emph{partial match} ($\geq 97\%$ identity), where the sequences differ only by minor isoform or annotation variation that is unlikely to alter the homology conclusion; and a \emph{mismatch} ($<97\%$ identity), where the agent selected a substantively different sequence from the human reference. We define a false negative as a case where the agent rejected a candidate that matched the human's selection.

\paragraph{Results.}
The agent reviewed 135 sequences across the 22 benchmark cases. Setting aside 11 sequences for which it found no valid protein in any database (9 correct rejections, 2 false negatives), the remaining 124 yielded 86 exact matches ($100\%$ identity; $69.4\%$), 21 partial matches ($\geq 97\%$; $16.9\%$), and 17 mismatches ($<97\%$; $13.7\%$).

Both false negatives trace to the same protein retrieved from two distinct databases; the agent anchored to a substantially longer incorrect human isoform, causing both candidates to be discarded by the deterministic $\pm$20\% length pre-filter.

The 17 mismatches arise from three complementary causes. First, in several cases the agent's reasoning correctly identified a candidate as a likely incorrect isoform yet still accepted it, a failure mode addressable through prompt tuning and tool-schema refinement. Second, the human analysts drew on specialized databases not yet accessible to the agent, yielding isoforms the agent could not reach, a limitation mitigable by extending the database catalog. Third, a subset of cases required non-standard expert procedures the agent is not currently equipped to replicate, such as excising a long C-terminal extension absent from the human reference and running a targeted similarity search on the remainder to recover the correct ortholog. The first two are tractable engineering gaps addressable through prompt tuning, tool-schema refinement, and an extended database catalog; the third remains an open research problem flagged for human review in deployment.

\subsubsection{HITL Refinement Regression Test}
\label{sec:eval:hitl}

Whereas \cref{sec:eval:seqhom} treats the human contribution as a frozen reference, the refinement loop itself (\cref{sec:method:loop}) places an expert in active control of the system. The question we ask here is: when an expert injects a targeted question to improve a specific section, does the agent produce a precise improvement, or does it introduce collateral damage elsewhere in the document~\citep{laban2026llms}?

We diff each report section pre/post a single HITL intervention at the sentence level. A domain expert then labels every changed sentence as \textbf{B}~(beneficial: correct new information or removal of incorrect content), \textbf{N}~(neutral: rephrasing without semantic change), or \textbf{H}~(harmful: factual error, hallucination, or loss of correct information). From these labels we define \emph{net improvement} $= (\text{B} - \text{H})\,/\,(\text{B} + \text{N} + \text{H})$ as the primary metric, alongside benefit rate~(B\,/\,total), harm rate~(H\,/\,total), and response rate~(questions addressed\,/\,total questions).

Twelve target-specific refinement requests are designed for a selected list of targets (\cref{fig:hitl_questions}), spanning the question dimensions of accuracy, completeness, prioritization, and consistency. Each question is delivered as a single intervention on an independent replicate, so that its effect is measured in isolation from other refinements.
We adopt a single-reviewer protocol because the B/N/H labels are anchored to verifiable criteria rather than subjective judgment: beneficial requires the added or modified content to be factually correct against publicly available data sources (genetic databases, expression atlases, clinical trial registries), harmful requires the introduction of a factual error or loss of correct information, and neutral is restricted to rephrasing without semantic change. This fact-checking structure, combined with the domain expertise required for verification, makes the labeling task suitable for a single qualified reviewer.

\begin{table}[htbp]
  \centering
  \caption{%
    HITL refinement outcomes stratified by question type (top) and report section (bottom). $n$: total changed sentences per stratum.
  }
  \label{tab:hitl_metrics}
  \small
  \setlength{\tabcolsep}{3pt}
  \begin{tabular}{@{} l c cccc @{}}
  \toprule
  \textbf{Stratum}
    & $n$ & \textbf{Benefit} & \textbf{Neutral} & \textbf{Harm} & \textbf{Net} \\
  \midrule
  \multicolumn{6}{@{}l}{\textit{By question type}} \\
  \quad Accuracy        &  20 & $85.0\%$ & $15.0\%$ & $\phantom{0}0.0\%$ & $+85.0\%$ \\
  \quad Completeness    & 274 & $79.9\%$ & $17.5\%$ & $\phantom{0}2.6\%$ & $+77.4\%$ \\
  \quad Consistency     &  63 & $47.6\%$ & $38.1\%$ & $14.3\%$ & $+33.3\%$ \\
  \quad Prioritization  & 113 & $66.4\%$ & $25.7\%$ & $\phantom{0}8.0\%$ & $+58.4\%$ \\
  \arrayrulecolor{sepgray}\midrule\arrayrulecolor{black}
  \multicolumn{6}{@{}l}{\textit{By report section}} \\
  \quad Clinical Safety      & 101 & $82.2\%$ & $15.8\%$ & $\phantom{0}2.0\%$ & $+80.2\%$ \\
  \quad Expression Profile   & 138 & $64.5\%$ & $28.3\%$ & $\phantom{0}7.2\%$ & $+57.2\%$ \\
  \quad Genetic Evidence     & 135 & $82.2\%$ & $14.8\%$ & $\phantom{0}3.0\%$ & $+79.3\%$ \\
  \quad Risk Assessment      &  96 & $60.4\%$ & $30.2\%$ & $\phantom{0}9.4\%$ & $+51.0\%$ \\
  \arrayrulecolor{sepgray}\midrule\arrayrulecolor{black}
  \rowcolor{rowgray}
  \textbf{Overall} & 470 & $72.6\%$ & $22.1\%$ & $\phantom{0}5.3\%$ & $+67.2\%$ \\
  \bottomrule
  \end{tabular}
\end{table}

\paragraph{Results.} 
Across $n=470$ changed sentences over 12 interventions, the refinement loop produced $+67.2\%$ net improvement ($72.6\%$~B, $22.1\%$~N, $5.3\%$~H), with every intervention triggering at least one change (\cref{tab:hitl_metrics}).

\emph{Completeness} ($n=274$, $+77.4\%$~net, $2.6\%$~H) dominates the changes, indicating that the refinement loop adds new beneficial content far more often than it fabricates. \emph{Prioritization} ($+58.4\%$~net, $8.0\%$~H) shows higher harm when re-ranking demotes content the reviewer judged equally salient. \emph{Consistency} is the only category with non-trivial harm ($14.3\%$~H, $+33.3\%$~net): asking the agent to reconcile statements across sections is meaningfully harder than retrieving, locally re-ranking, or reorganizing content already in scope. \emph{Accuracy} was probed by only a single intervention ($n=20$ changed sentences, $+85.0\%$~net) and is reported as suggestive rather than conclusive.

\emph{Clinical Safety} ($+80.2\%$) and \emph{Genetic Evidence} ($+79.3\%$) refine cleanly, while \emph{Risk Assessment} ($+51.0\%$~net, $9.4\%$~H) is unsurprisingly the noisiest, since it is the only section whose content is integrative rather than evidence-local.

\section{Discussion}
\label{sec:discussion}
\tool is designed to explore whether a hierarchical multi-agent architecture with human-in-the-loop design can accelerate \tsa report drafting under expert oversight.
Across the four evaluations, \tool produces reproducible endpoint-level claims (\cref{sec:eval:concordance}), achieves high sentence-level grounding under subagent decomposition with cross-section memory (\cref{sec:eval:esr}), shows substantial agreement with a human reference on the ortholog benchmark (\cref{sec:eval:seqhom}), and yields net-positive expert-driven refinement (\cref{sec:eval:hitl}). 
Anticipated challenges in deploying systems of this kind include sensitivity to ambiguous or underspecified target queries, variability in evidence prioritization across sections, reconciliation of conflicting evidence across databases, and adaptation to targets with sparse literature coverage. \tool incorporates retrieval constraints, mandatory citation enforcement, and embedded human validation checkpoints intended to address these. Future work will further explore automatic evidence conflict resolution, tighter integration with internal proprietary databases, and extension to related early-safety workflows such as genotoxicity and cardiotoxicity assessment.

More broadly, target safety assessment is a resource-intensive step in drug discovery, and \tool is, to our knowledge, among the first frameworks specifically tailored to the needs of toxicologists and pharmaceutical scientists. By reducing the mechanical burden of evidence synthesis and report drafting while preserving expert oversight through section-level refinement and progressive personalization, \tool aims to contribute to a broader vision of AI-augmented pharmaceutical safety science where agentic AI and human expertise are complementary.

\section*{Author Contributions}
Doktorova initiated, directed, and supervised the project; Doktorova, Guerard, Hatje, Jiang, and Zheng formulated the problem; Zheng conceived the framework, designed the system architecture and evaluation methodology, and implemented the entire system; Zheng, Jiang, Tokar, and Cheng designed the evaluation metrics; Zheng, Tokar, and Cheng conducted the experiments; Zheng, Jiang, Tokar, Cheng, and Serra performed the analysis; Doktorova, Guerard, Hatje, and Jiang translated domain expertise into tool design specifications; Doktorova, Zheng, Jiang, and Hatje curated the data sources and provided data-access resources; Jiang, Zheng, and Doktorova wrote the manuscript; all authors provided expert validation, contributed to user requirements and quality control, participated in discussions, and revised the manuscript.

\section*{Acknowledgements}

The authors thank colleagues across Roche for their contributions, including technical input, scientific discussions, and feedback on system design and evaluation. Platform and agentic workflow development were carried out within Pharmaceutical Sciences (Predictive Modelling) and the Computational Sciences Center of Excellence (Data and Digital Catalysts, Computational Biology and Medicine). Scientific contributions, including target safety assessment expertise and validation, were provided by Pharmaceutical Sciences (Predictive Modelling, Translational Safety) and the Computational Sciences Center of Excellence (Computational Biology and Medicine). 
The authors also acknowledge support in data engineering, infrastructure, and tool integration that enabled the development and deployment of the system.

\bibliography{ref}
\bibliographystyle{icml2026}

\newpage
\appendix
\onecolumn

\section{Data Sources}
\label{app:datasources}

The data sources backing the tool interfaces span the five lines of evidence described in \cref{sec:bg:tsa}.
PubMed \citep{white2020pubmed} provides primary literature evidence across all sections. Other sources are domain specific, spanning genetics, genomics, transcriptomics, target homology, pharmacological and clinical evidence.
We use Ensembl~\citep{birney2004ensembl, cunningham2022ensembl} as primary source to get genomic coordinates, transcript and protein annotations, cross-species orthologues, protein domain information, germline and somatic variants with phenotype associations, and regulatory feature annotations.
NCBI RefSeq~\citep{o2016reference} complements this with curated reference sequences for genes and transcripts essential for standardized genomic annotation.
UniProt~\citep{uniprot2023uniprot} provides curated protein sequences, functional annotations, and tissue expression data essential for evaluating both on-target and off-target safety liabilities. 

Functional characterization is further supported by Gene Ontology (GO)~\citep{ashburner2000gene, gene2023gene}, which supplies standardized molecular functions, biological processes and cellular components, enabling systematic characterization of target biology and potential mechanism-based toxicities.
Pathway-level context is provided by KEGG~\citep{kanehisa2000kegg, kanehisa2023kegg} and Reactome~\citep{gillespie2022reactome}, which map targets to metabolic, signaling, and disease pathways critical for understanding downstream physiological consequences of target modulation.

Human genetic and clinical evidence is integrated from multiple sources.
The NHGRI-EBI GWAS Catalog \citep{buniello2019nhgri} aggregates genome-wide association signals linking variants to phenotypes.
ClinVar \citep{landrum2018clinvar} supplies curated variant--disease interpretations including pathogenicity classifications.
The Human Protein Atlas \citep{uhlen2010towards, karlsson2021single} provides tissue- and cell-type-level protein and RNA expression profiles.
GTEx \citep{gtex2020gtex} supplies tissue-specific expression and eQTL data across human tissues.
The EMBL-EBI Expression Atlas \citep{papatheodorou2020expression} offers curated RNA-seq expression profiles across diseases, cell types, and developmental stages. 

To explore translatability, we use Open Targets~\citep{ochoa2021open, ochoa2023next}, which integrates genetic, genomic, and clinical evidence, to validate disease--target associations and extract reported adverse events from clinical studies and approved drugs.
The Mouse Genome Informatics database (MGI) \citep{baldarelli2024mouse} provides phenotype ontology annotations, knockout phenotype data and human-mouse disease connections to facilitate exploration of candidate genes and investigation of phenotypic similarity between mouse models and human patients.
ClinicalTrials.gov\footnote{\url{https://clinicaltrials.gov}} supplies clinical study records including safety endpoints and adverse event reports from interventional trials.
AdisInsight\footnote{\url{https://adisinsight.springer.com}} provides competitive intelligence, including known severe adverse events associated with drugs modulating the same target.


\section{Claim-Level Self-Consistency: Full Metric Definition}
\label{app:concordance_method}

This appendix provides the complete metric definition summarized in \cref{sec:eval:concordance}.

\textit{Step~1: Claim extraction.}
Each safety endpoint's finding text is decomposed into atomic claims by an LLM with toxicology-informed system prompt.
Each claim is a structured pair $c = (\textit{observation}, \textit{species})$, where \textit{species} encodes the biological context in which the observation was made.

\textit{Step~2: Claim pair scoring.}
For two claims $c_1$ and $c_2$ from the two runs being compared, we define:

\begin{equation}
\text{score}(c_1, c_2) = \text{sim}(c_1.\textit{obs},\; c_2.\textit{obs}) \cdot \phi_s(c_1, c_2)
\end{equation}

where $\text{sim}(\cdot, \cdot) \in \{0, 1\}$ is an LLM-as-judge binary similarity score~\citep{zheng2023judging, liu2023geval, jiang2025clear}, which is scaled by $\phi_s$, a species context matching function, encoding domain-specific constraints on biological translatability:

\begin{equation}
\phi_s(c_1, c_2) = \begin{cases}
  1.0 & \text{same species,} \\
  0.75 & \text{same species group,} \\
  0.5 & \text{different species group,} \\
  0 & \text{otherwise.}
\end{cases}
\end{equation}

The four-level weighting reflects regulatory toxicology practice on inter-species translatability~\citep{ich2011s6}; species-group definitions are in \cref{tab:species_groups} (\cref{app:eval_defs}). Contradictory claims (\eg ``dose-related neutropenia'' \vs ``no significant hematotoxicity'') are explicitly scored as zero by the LLM judge.
Same-species matches receive full credit ($\phi_s = 1.0$): hepatic fibrosis observed in two independent rat studies is the same evidence type. Same-group matches receive partial credit ($\phi_s = 0.75$), as inter-species differences within a group indicate that findings are convergent but not equivalent. Cross-group matches receive reduced credit ($\phi_s = 0.5$): the same phenotype observed in rodent preclinical studies and in human clinical trials represents qualitatively different evidence, as animal findings are not always predictive of human outcomes due to inter-species differences in gene expression, target abundance, receptor functionality, signaling pathways, metabolism, tissue distribution, and compensatory physiological mechanisms.

\textit{Step~3: Symmetric optimal matching.}
We adopt the greedy optimal matching formulation from BERTScore~\citep{zhang2020bertscore}, which computes precision and recall by independently finding the best match for each element, and combine it with SAFE's F1 aggregation:

\begin{align}
\text{Recall} &= \frac{1}{|C_1|} \sum_{c \in C_1} \max_{c' \in C_2} \text{score}(c, c'), \\[4pt]
\text{Precision} &= \frac{1}{|C_2|} \sum_{c' \in C_2} \max_{c \in C_1} \text{score}(c', c), \\[4pt]
\text{Claim-F1} &= \frac{2 \cdot \text{Precision} \cdot \text{Recall}}{\text{Precision} + \text{Recall}}.
\end{align}

The symmetric formulation distinguishes our approach from FActScore and SAFE, which operate unidirectionally (generated $\to$ reference).
In our setting, neither run is assumed to be definitively complete: one run may identify findings the other omits.
The greedy matching permits many-to-one alignments, accommodating cases where one run expresses a single comprehensive finding that the other decomposes into multiple granular claims.

\newpage


\section{Claim-Level Self-Consistency: Endpoint and Species Definitions}
\label{app:eval_defs}

\begin{table}[h]
\centering
\caption{Fifteen standardized safety endpoints used in the integrated risk assessment, aligned to MedDRA System Organ Classes~\citep{brown1999medical}.}
\label{tab:endpoints}
\begin{tabular}{@{}cl@{}}
\toprule
\textbf{\#} & \textbf{Safety Endpoint} \\
\midrule
1  & Hepatotoxicity \\
2  & Renal toxicity \\
3  & Bone marrow toxicity / hematotoxicity \\
4  & Neurological / CNS / neurodevelopmental toxicity \\
5  & Cardiovascular toxicity \\
6  & Reproductive / developmental toxicity \\
7  & Ocular toxicity \\
8  & Inflammatory syndrome \\
9  & Immune dysfunction \\
10 & Metabolic dysregulation \\
11 & Skin and subcutaneous tissue disorders \\
12 & Gastrointestinal toxicity \\
13 & Pulmonary toxicity \\
14 & Endocrine disruption \\
15 & Pregnancy / puerperium / perinatal conditions \\
\bottomrule
\end{tabular}
\end{table}

\begin{table}[h]
\centering
\caption{Species group definitions for the species context matching function $\phi_s$. Non-human \textit{in vitro} systems (\eg hERG assays, rodent hepatocytes, genotoxicity panels) are classified as experimental modalities and excluded from species-based matching.}
\label{tab:species_groups}
\begin{tabular}{@{}lp{8cm}@{}}
\toprule
\textbf{Species Group} & \textbf{Members} \\
\midrule
Rodent & Mouse, rat, hamster, guinea pig \\
Non-rodent & NHP (cynomolgus macaque, marmoset), dog, minipig, rabbit \\
Human & Human genetic evidence, clinical adverse event data, human-derived \textit{in vitro} systems (iPSC-derived cells, primary cells) \\
Alternative screening models & Zebrafish embryo \\
\bottomrule
\end{tabular}
\end{table}


\section{Claim-Level Self-Consistency: Prompt Specifications}
\begin{tcolorbox}[
  title=Self-Consistency Evaluation Instructions,
  enhanced,
  breakable,
  colback=white,
  colframe=black!60,
  boxrule=0.6pt,
  arc=2pt,
  left=1.5mm,
  right=1.5mm,
  top=1mm,
  bottom=1mm
]
\begin{Verbatim}[fontsize=\small,breaklines=true,breakanywhere=true,breaksymbolleft=,breaksymbolright=]
You are a toxicology expert. Determine whether these two observations describe the same safety finding.

Observation A: "{text_a}"
Observation B: "{text_b}"

Score EXACTLY one of:
- 1 = Match: The observations describe the same finding if ANY of these rules apply:
 - "synonym": Synonym or paraphrase. Different wording for the same concept, for example, "liver fibrosis" = "hepatic fibrosis"
 - "cause_consequence": Cause and consequence within the same organ/system. One is the direct cause or result of the other, both involving the same organ system, for example, "reduced K+ currents" = "prolonged QT interval" (both cardiac electrophysiology); "VE-cadherin disruption" = "vascular hemorrhage" (both vascular) 
 - "mechanism_outcome": Mechanism and clinical outcome within the same system. A molecular or cellular mechanism paired with its clinical manifestation in the same system, for example, "demyelination" = "nerve conduction slowing" (both peripheral nervous system)
 - "biomarker_disease": Biomarker evidence and the condition it indicates. A measurable marker and the disease or toxicity it reflects, such as "elevated ALT" = "hepatotoxicity"; "proteinuria" = "podocyte injury".
 - "includes": One includes the other. One observation is a specific instance, subtype, or manifestation of the other. This happens when one is vague and the other is specific, for example, if one term is a broad category that encompasses the other. This includes "fulminant hepatic failure" = "hepatotoxicity", "ventricular tachycardia" = "cardiac arrhythmia", "skin reactions" = "acneiform rash" (general term includes the specific). 
 - "same_phenotype": Same underlying phenotype observed across different experimental setups or with different measurements. Both findings are phenotypic consequences of modulating the same target within the same organ or system and belong to the toxicological profile of the same target perturbation. For example, "hepatic steatosis in constitutive knockout" = "hepatic steatosis with antibody treatment" (same phenotype confirmed across genetic and pharmacological models).
 - "same_syndrome": Both findings are recognized features of the same syndrome with the same clinical grouping. For example, "delayed puberty" = "premature follicle depletion" (both are recognized features of premature ovarian insufficiency).
 - "same_pathway": Different molecules in the same biological pathway. Both observations describe changes in molecules that belong to the same biological signaling pathway. For example, "elevated IL-12p40" = "elevated IL-1alpha" (both pro-inflammatory cytokines), "reduced DOPAC levels" = "reduced striatal dopamine" (both dopaminergic pathway).  
 - "same_pathology": Same pathology finding with different quantitative description. Same underlying pathology differing only in time course, frequency, severity or grade is a MATCH. This includes "Grade 1 thrombocytopenia" = "grade 3-4 thrombocytopenia"; "Acute liver injury" = "chronic hepatotoxicity".



- 0 = No match: The observations are different findings if EITHER applies:
 - "different_organs": The findings affect different organs or body systems, even if caused by the same gene, drug, or pathway.    For example, "B cell depletion" does not match "pericarditis" (immune vs cardiac); "reduced fertility" does not match "reduced hypothalamic neurons" (reproductive vs CNS); "headache" does not match "elevated IFN-alpha" (neurological symptom vs systemic cytokine).
 - "contradictory": The observations describe opposite directions of the same parameter or mutually exclusive states. This includes the cases where "OPC proliferation" does not match "OPC depletion", "protection from neurodegeneration" does not match "neurodegeneration"; "no hepatotoxicity observed" does not match "hepatotoxicity". 
 - "distinct": The observations describe different findings or aspects, even when referring to the same organ. They do not contradict each other, but they also do not correspond to the same observation.


Species context: Species weighting is handled separately and not considered when analyzing the finding itself.
First classify which rule applies, then assign the score.
First go through ALL categories in the MATCH cases, and move to the NO MATCH category ONLY when NONE of the MATCH rules apply. 

Categories: synonym, cause_consequence, mechanism_outcome, biomarker_disease, includes, same_phenotype, same_syndrome, same_pathway, same_pathology, different_organs, contradictory, distinct.


Return ONLY a JSON object: {{"rule": "<category>", "score": <int>}}

\end{Verbatim}
\end{tcolorbox}
\captionsetup{hypcap=false}
\captionof{figure}{Full instruction prompt used for the self-consistency evaluation.}
\label{fig:judge_prompt_full}

\begin{tcolorbox}[
  title=Claim Extraction Instructions,
  enhanced,
  breakable,
  colback=white,
  colframe=black!60,
  boxrule=0.6pt,
  arc=2pt,
  left=1.5mm,
  right=1.5mm,
  top=1mm,
  bottom=1mm
]
\begin{Verbatim}[fontsize=\small,breaklines=true,breakanywhere=true,breaksymbolleft=,breaksymbolright=]
You are a pharmacology/toxicology expert. Given a toxicity finding from a Target Safety Assessment report, decompose it into atomic claims.

Each claim must be a self-contained observation linked to its species.

Toxicity Endpoint: {endpoint_name}
Target: {target}
Finding text:
\"\"\"{finding}\"\"\"

Decompose into a JSON list of claims:

{{
  "claims": [
    {{
      "observation": "specific phenotype or finding as a short phrase",
      "species": "one of: human, mouse, rat, hamster, guinea_pig, dog, minipig, rabbit, non_human_primate, zebrafish"
    }}
  ]
}}

Rules:
- Each claim = ONE distinct observation tied to ONE species
- Only extract POSITIVE findings (observed phenotypes, toxicities, or adverse effects). Do NOT extract negative statements such as "no adverse findings", "no toxicity observed", "not reported", "no dose-limiting findings identified", or any statement indicating absence of effect.
- Extract only homozygous phenotypes in animals observed after knock out (0% function of the target of interest) (e.g., "increased adipose tissue mass upon Inhbe overexpression", "hepatotoxicity in KO mice")
- Do NOT extract gene/protein expression patterns (e.g., "target X is expressed in tissue Y", "high expression in brain"), tissue distribution data, or mechanistic descriptions that are not observed adverse phenotypes. However, DO extract phenotypic consequences of overexpression/knockout experiments (e.g., "increased adipose tissue mass upon Inhbe overexpression", "hepatotoxicity in KO mice") — these are valid toxicology findings even though the experimental model involves gene manipulation.
- Do NOT extract findings from non-human in vitro systems (e.g., hERG assays, rodent hepatocytes, genotoxicity panels, cell line experiments). These are experimental modalities, not in vivo toxicology findings. Only extract in vivo observations or human clinical/genetic data.
- If a sentence mentions multiple phenotypes, split into separate claims
- "observation": be specific (e.g., "aortic aneurysm formation" not "cardiovascular effects")
- "species": the species in which this was observed, classified per ICH S6 regulatory toxicology categories:
  - rodent: mouse, rat, hamster, guinea_pig
  - non-rodent: non_human_primate (cynomolgus macaque, marmoset), dog, minipig, rabbit
  - human: encompasses human genetic evidence, clinical adverse event data, and human-derived in vitro systems (e.g., iPSC-derived cells, primary human hepatocytes)
  - alternative screening models: zebrafish
  - Do NOT use "other". If a finding does not fit the above species categories, skip it entirely.

Return ONLY valid JSON, no markdown fencing.


SPECIES_ALIASES = {
    "mice": "mouse", "murine": "mouse", "mus musculus": "mouse",
    "rats": "rat", "rattus": "rat",
    "hamsters": "hamster", "syrian hamster": "hamster",
    "guinea pig": "guinea_pig", "guinea pigs": "guinea_pig", "cavia": "guinea_pig",
    "dogs": "dog", "beagle": "dog", "canine": "dog",
    "minipigs": "minipig", "mini-pig": "minipig", "pig": "minipig", "swine": "minipig", "porcine": "minipig",
    "rabbits": "rabbit", "lapine": "rabbit",
    "nhp": "non_human_primate", "monkey": "non_human_primate", "monkeys": "non_human_primate",
    "primate": "non_human_primate", "primates": "non_human_primate",
    "cynomolgus": "non_human_primate", "macaque": "non_human_primate", "rhesus": "non_human_primate",
    "marmoset": "non_human_primate",
    "zebrafish": "zebrafish", "danio rerio": "zebrafish", "danio": "zebrafish",
    "humans": "human", "patients": "human", "clinical": "human",
}

VALID_SPECIES = {"human", "mouse", "rat", "hamster", "guinea_pig", "dog", "minipig",
                 "rabbit", "non_human_primate", "zebrafish", "other"}
\end{Verbatim}
\end{tcolorbox}
\captionsetup{hypcap=false}%
\captionof{figure}{Full instruction prompt used for extracting the atomic claims from the safety endpoint.}
\label{fig:tox_endpoint_prompt}


\section{Claim-Level Self-Consistency: Claim Comparison Statistics }

\Cref{tab:species_relation_rules} provides a full breakdown of all the claim pairs generated during the self-consistency evaluation described in \cref{sec:eval:concordance}, stratified by species relation and scoring rule. Each pair is scored using the two-factor scheme of \cref{fig:judge_prompt_full}: a rule-based scorer first classifies the semantic relationship between the two claims (Match rules contribute a positive score; No Match rules yield zero), and the resulting score is then scaled by the species weight $\phi_s \in \{0.5, 0.75, 1.0\}$ reflecting the biological translatability of the pair.

The majority of pairs fall in the \textit{Same Species} category (67.4\%), consistent with the expectation that independent runs on the same target should predominantly surface evidence from the same biological contexts.

Within this category, the dominant No Match rule is \textit{Different Organs} (25.48\% of Same Species pairs), which arises when two claims describe the same species but distinct anatomical systems and thus cannot be considered concordant.
Cross-species pairs (\textit{Diff.\ Group}, 31.7\%) show a higher concentration of \textit{Mechanism Outcome} matches (21.23\%), where one claim describes a molecular or cellular mechanism and the other its clinical manifestation within the same organ system.
This pattern is elevated in cross-species comparisons because preclinical runs tend to capture mechanistic observations (\eg hepatocyte apoptosis) while independent runs drawing on clinical literature surface the downstream adverse manifestation (\eg drug-induced liver injury), both localized to the same system but attributed to different species groups.

\begin{table}[htbp]
  \centering
\caption{%
    Distribution of claim-pair scoring rules stratified by species relation, derived from  35 targets and 15 safety and toxicological safety endpoints. For each target, 10 independent LLM runs were executed under identical prompts, producing $\binom{10}{2}=45$ run pairs per target. Within each run pair, every claim extracted by run~$a$ is cross-compared against every claim extracted by run~$b$ for the same endpoint.
    Each pair is assigned a rule by a rule-based scorer in ~\cref{fig:judge_prompt_full}.
    The \textit{species relation} field captures whether the two claims in a pair reference the same biological species (\textit{Same Species}, 67.4\%), species from different groups (\textit{Diff.\ Group}, 31.7\%), or species from the same group but distinct entries (\textit{Same Group}, 0.9\%).
    \textit{\% within}: fraction of claim pairs within that species relation category assigned
    the given rule. \textit{\% of total}: fraction across claim pairs.
  }
  \label{tab:species_relation_rules}
  \small
  \begin{tabular}{@{} l l  rr  rr  rr @{}}
  \toprule
  & &
    \multicolumn{2}{c}{\textbf{Same Species}} &
    \multicolumn{2}{c}{\textbf{Diff.\ Group}} &
    \multicolumn{2}{c}{\textbf{Same Group}} \\[2pt]
  & &
    \multicolumn{2}{c}{\small 67.4\% of all records} &
    \multicolumn{2}{c}{\small 31.7\% of all records} &
    \multicolumn{2}{c}{\small 0.9\% of all records} \\
  \cmidrule(lr){3-4}\cmidrule(lr){5-6}\cmidrule(lr){7-8}
  \textbf{Category} & \textbf{Rule}
    & \% within & \% of total
    & \% within & \% of total
    & \% within & \% of total \\
  \midrule

  \multirow{9}{*}{\textbf{Match}}
    & \texttt{biomarker\_disease}  & $1.91\%$ & $1.29\%$ & $2.04\%$ & $0.65\%$ & $2.53\%$ & $0.02\%$ \\
    & \texttt{cause\_consequence}  & $15.73\%$ & $10.60\%$ & $9.16\%$ & $2.90\%$ & $3.08\%$ & $0.03\%$ \\
    & \texttt{mechanism\_outcome}  & $10.45\%$ & $7.05\%$ & $21.23\%$ & $6.73\%$ & $4.95\%$ & $0.04\%$ \\
    & \texttt{same\_pathology}     & $7.90\%$ & $5.33\%$ & $1.50\%$ & $0.47\%$ & $2.64\%$ & $0.02\%$ \\
    & \texttt{same\_pathway}       & $4.85\%$ & $3.27\%$ & $4.03\%$ & $1.28\%$ & $0.55\%$ & $0.00\%$ \\
    & \texttt{same\_phenotype}     & $8.34\%$ & $5.62\%$ & $13.37\%$ & $4.24\%$ & $12.54\%$ & $0.11\%$ \\
    & \texttt{same\_syndrome}      & $7.58\%$ & $5.11\%$ & $4.84\%$ & $1.53\%$ & $0.22\%$ & $0.00\%$ \\
    & \texttt{synonym}            & $4.47\%$ & $3.01\%$ & $0.15\%$ & $0.05\%$ & $3.41\%$ & $0.03\%$ \\
    & \texttt{includes}           & $8.96\%$ & $6.04\%$ & $8.44\%$ & $2.67\%$ & $7.37\%$ & $0.07\%$ \\

  \arrayrulecolor{sepgray}\midrule\arrayrulecolor{black}

  \rowcolor{rowgray}
    & \texttt{contradictory}      & $0.97\%$ & $0.65\%$ & $1.25\%$ & $0.40\%$ & $2.86\%$ & $0.03\%$ \\
  \rowcolor{rowgray}
    & \texttt{different\_organs}   & $25.48\%$ & $17.18\%$ & $29.66\%$ & $9.40\%$ & $49.61\%$ & $0.44\%$ \\
  \rowcolor{rowgray}
  \multirow{-3}{*}{\textbf{No Match}}
    & \texttt{distinct}          & $3.37\%$ & $2.27\%$ & $4.33\%$ & $1.37\%$ & $10.23\%$ & $0.09\%$ \\
  
  \bottomrule
  \end{tabular}
  \end{table}


\section{Sequence Homology Procedure}
\label{app:seqhom_procedure}

This appendix provides the full procedure summarized in \cref{sec:eval:seqhom}.

The agent carries out the core of the ortholog identification task in four main steps:
\begin{enumerate}
\item Query the ortholog database for each preclinical species.
\item Review the ortholog candidates returned and retain those passing the deterministic pre-filter (length deviation $\leq 20\%$, pairwise mismatch $\leq 60\%$).
\item For any species with no valid candidate from step 2, run a targeted sequence similarity search against the human canonical sequence.
\item Review the search results and select the highest-confidence candidate that passes the pre-filter.
\end{enumerate}

When no suitable candidate can be found in either pass, the agent excludes the species from the analysis and reports every database it searched together with the rejection reason for each candidate. This allows the user to exercise their own judgment by accepting the exclusion, manually selecting a borderline candidate, or invoking a non-standard procedure (\eg excising a divergent C-terminal extension and re-running the similarity search on the remainder).


\section{HITL: Target-Specific Refinement Questions}
\label{app:hitl_questions}
\begin{tcolorbox}[
  title=Target A (inhibitor),
  enhanced,
  breakable,
  colback=white,
  colframe=black!60,
  boxrule=0.6pt,
  arc=2pt,
  left=1.5mm,
  right=1.5mm,
  top=1mm,
  bottom=1mm
]
\begin{Verbatim}[fontsize=\small,breaklines=true,breakanywhere=true,breaksymbolleft=,breaksymbolright=]
HITL_A_Q1 (Clinical Safety, Completeness):
"Revise this section to include any clinical-stage compounds targeting this target or the oxysterol pathway, and add available safety data for these compounds."

HITL_A_Q2 (Risk Assessment, Completeness):
"This target is involved in both immune cell migration and lipid metabolism. Update the integrated assessment to ensure both axes of risk are explicitly reflected and discussed."

HITL_A_Q3 (Genetic Evidence, Completeness):
"Knockout mice show altered germinal center formation. Ensure this phenotype is included in the section and add a discussion of its relevance to immunosuppression risk."

HITL_A_Q4 (Genetic Evidence, Accuracy):
"Update this section to include any reported loss-of-function variants from population databases (e.g., gnomAD), with correctly described associated phenotypes."
\end{Verbatim}
\end{tcolorbox}

\vspace{0.5em}

\begin{tcolorbox}[
  title=Target B (activator),
  enhanced,
  breakable,
  colback=white,
  colframe=black!60,
  boxrule=0.6pt,
  arc=2pt,
  left=1.5mm,
  right=1.5mm,
  top=1mm,
  bottom=1mm
]
\begin{Verbatim}[fontsize=\small,breaklines=true,breakanywhere=true,breaksymbolleft=,breaksymbolright=]
HITL_B_Q1 (Expression Profile, Completeness):
"This target is predominantly neuronal. Update the expression analysis to include any non-neuronal tissues with detectable expression and discuss potential off-target safety concerns."

HITL_B_Q2 (Genetic Evidence, Completeness):
"Update this section to include TDP-43 related variants that affect target expression, and discuss whether these inform the safety profile of target activation."

HITL_B_Q3 (Clinical Safety, Completeness):
"Update this section to cover any clinical programs targeting this protein or the TDP-43 axis. If none exist, state this explicitly and discuss the implications for safety assessment."

HITL_B_Q4 (Risk Assessment, Prioritization):
"Given that this is an activator with neuronal-restricted expression, revise the risk assessment to clearly state the primary safety concern as the top-ranked risk."
\end{Verbatim}
\end{tcolorbox}

\vspace{0.5em}

\begin{tcolorbox}[
  title=Target C (inhibitor),
  enhanced,
  breakable,
  colback=white,
  colframe=black!60,
  boxrule=0.6pt,
  arc=2pt,
  left=1.5mm,
  right=1.5mm,
  top=1mm,
  bottom=1mm
]
\begin{Verbatim}[fontsize=\small,breaklines=true,breakanywhere=true,breaksymbolleft=,breaksymbolright=]
HITL_C_Q1 (Genetic Evidence, Completeness):
"Target deficiency causes a lysosomal storage disease. Update this section to include key phenotypes of knockout models and add a discussion of their relevance to dose-dependent pharmacological inhibition."

HITL_C_Q2 (Expression Profile, Prioritization):
"This target is a lysosomal enzyme with broad tissue expression. Revise the expression analysis to prioritize which organs are most vulnerable to substrate accumulation upon inhibition."

HITL_C_Q3 (Clinical Safety, Completeness):
"Update this section to include any clinical compounds targeting this enzyme or the ceramide pathway, along with their reported adverse event profiles."

HITL_C_Q4 (Risk Assessment, Consistency):
"Revise the risk assessment to explicitly distinguish between complete loss-of-function (Y disease) and partial pharmacological inhibition, and add a discussion of the therapeutic window."
\end{Verbatim}
\end{tcolorbox}
\captionsetup{type=figure,hypcap=false}%
\captionof{figure}{Representative target-specific refinement questions used in the HITL regression test (\cref{sec:eval:hitl}). Each question is designed by a safety scientist to probe a specific report section along one of four dimensions: accuracy, completeness, prioritization, or consistency. Target names are anonymized. The figure shows all twelve interventions used in the regression test, spanning three representative targets with four questions each.}
\label{fig:hitl_questions}

\end{document}